\documentclass[runningheads]{llncs}

\usepackage{eccv}

\usepackage{eccvabbrv}
\usepackage{wrapfig}

\usepackage{graphicx}
\usepackage{booktabs}

\usepackage[accsupp]{axessibility}  

\makeatletter
\newcommand{\printfnsymbol}[1]{%
        \textsuperscript{\@fnsymbol{#1}}%
}
\makeatother

\usepackage{hyperref}

\usepackage{orcidlink}
\usepackage{multirow}

\definecolor{turquoise}{cmyk}{0.65,0,0.1,0.3}
\definecolor{purple}{rgb}{0.65,0,0.65}
\definecolor{dark_purple}{rgb}{0.5,0,0.5}
\definecolor{dark_green}{rgb}{0, 0.5, 0}
\definecolor{orange}{rgb}{0.8, 0.6, 0.2}
\definecolor{red}{rgb}{0.8, 0.2, 0.2}
\definecolor{darkred}{rgb}{0.6, 0.1, 0.05}
\definecolor{blueish}{rgb}{0.0, 0.3, .6}
\definecolor{light_gray}{rgb}{0.7, 0.7, .7}
\definecolor{pink}{rgb}{1, 0, 1}
\definecolor{greyblue}{rgb}{0.25, 0.25, 1}

\usepackage{algorithm}
\usepackage{listings}

\usepackage{etoolbox}
\makeatletter
\AfterEndEnvironment{algorithm}{\let\@algcomment\relax}
\AtEndEnvironment{algorithm}{\kern2pt\hrule\relax\vskip3pt\@algcomment}
\let\@algcomment\relax
\newcommand\algcomment[1]{\def\@algcomment{\footnotesize#1}}
\renewcommand\fs@ruled{\def\@fs@cfont{\bfseries}\let\@fs@capt\floatc@ruled
  \def\@fs@pre{\hrule height.8pt depth0pt \kern2pt}%
  \def\@fs@post{}%
  \def\@fs@mid{\kern2pt\hrule\kern2pt}%
  \let\@fs@iftopcapt\iftrue}
\makeatother

\newcommand{\qs}[1]{{\color{black}#1}}

\newcommand{\model}{\textsc{Forest2Seq}\xspace}

\usepackage{colortbl}

\begin{document}

\title{\model{}: Revitalizing Order Prior for Sequential Indoor Scene Synthesis} 

\titlerunning{\model}

\author{
Qi Sun\inst{1}\thanks{Equal contributions; Work carried out at SFU by Hang.
}
\and
Hang Zhou\inst{2}\printfnsymbol{1}
\and
Wengang Zhou\inst{1}
\and 
Li Li\inst{1}
\and
Houqiang Li\inst{1}
}

\authorrunning{Q. Sun and H. Zhou \etal}
\institute{USTC
\and 
Simon Fraser University
}

\maketitle

\begin{center}
    \centering
    \captionsetup{type=figure}
    \vspace{-2mm}
    \includegraphics[width=0.97\linewidth]{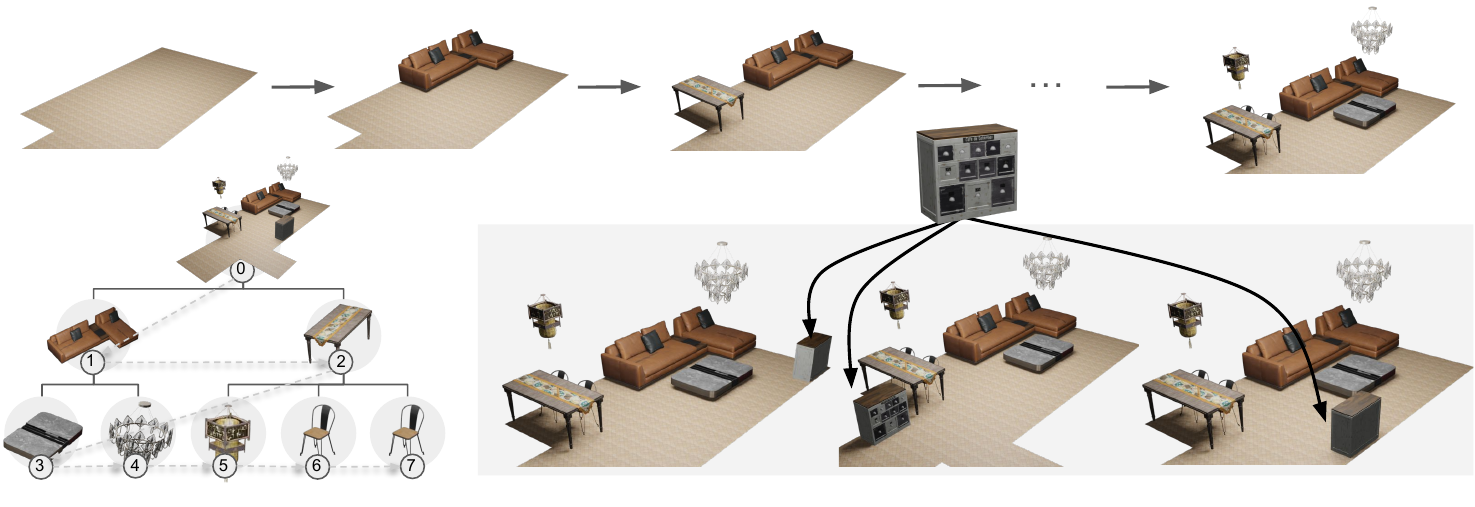}
    \captionof{figure}{We present \model that mines the implicit hierarchy from the scene (bottom left), employing the tree-derived ordering \qs{as significant prior} to direct the sequential indoor scene synthesis (top). The presence of \qs{placing}-adaptable furniture items (bottom right), exemplified by the cabinets, necessitate the evolution from a \qs{single} tree to scene forest representation.
    \vspace{-3mm}
    }
    \label{fig:teaser}
\end{center}%

\begin{abstract}
Synthesizing realistic 3D indoor scenes is a challenging task that traditionally relies on manual arrangement and annotation by expert designers. Recent advances in autoregressive models have automated this process, but they often lack semantic understanding of the relationships and hierarchies present in real-world scenes, yielding limited performance. In this paper, we propose \model, a framework that formulates indoor scene synthesis as an order-aware sequential learning problem. \model organizes the inherently unordered collection of scene objects into structured, ordered hierarchical scene trees and forests. By employing a clustering-based algorithm and a breadth-first traversal, \model derives meaningful orderings and utilizes a transformer to generate realistic 3D scenes autoregressively. Experimental results on standard benchmarks demonstrate \model's superiority in synthesizing more realistic scenes compared to top-performing baselines, with significant improvements in FID and KL scores. Our additional experiments for downstream tasks and ablation studies also confirm the importance of incorporating order as a prior in 3D scene generation.

\end{abstract}
\section{Introduction}
\label{sec:intro}

Creating realistic virtual 3D indoor scenes has long been an expensive, labor-intensive process~\cite{msl11, Siddhartha2013uist, yyt15}, requiring expert designers to manually arrange and annotate every piece to populate these rich environments~\cite{qi2018human, kar2019meta, devaranjan2020meta}. Recent advances in indoor scene synthesis, however, enable automating this process by generating plausible room layouts simply by taking high-level room type and layout~\cite{paschalidou2021atiss, tang2023diffuscene, wang2019planit, ritchie2019fast} as inputs. This has immense potential, enabling virtual product showcasing for retailers~\cite{ikea, Leimer10.1145/3550469.3555425}, automating environment creation for movies~\cite{koh2023simple}, games~\cite{Zhao2021Luminous} and complex visualization~\cite{patil2023advances}, and providing rich training data for 3D scene understanding AI models~\cite{Matterport3D, straub2019replica, procthor}.

Early indoor synthesis approaches were formulated as a prior constraint optimization task~\cite{msl11, fisher2012example, jls12, sch16}, achieving promising results. They encoded rules and priors about functional relationships between objects~\cite{yu2011make, yyt15} (like couches facing TVs) as well as human use-case constraints~\cite{ma2018language, zhao2016relationship}, which are time-consuming and skill-dependent. As a result, this hand-crafted prior approach lacks flexibility and generality as the 3D indoor scene becomes more complex~\cite{paschalidou2021atiss, ritchie2019fast}. 

As an alternative to hand-crafted priors, deep generative models have been employed to learn scene priors directly from data, without the need for manually specified rules or constraints. Such approaches include autoregressive transformer models~\cite{paschalidou2021atiss, wang2021sceneformer, cofs} and CNN-based methods~\cite{wang2019planit, weiss2018fast}. While autoregressive models generate objects in sequential order, a key limitation is that this order is arbitrary~\cite{vinyals2015order, eriguchi-etal-2016-tree, xu2018graph2seq} and lacks semantic understanding of the relationships and hierarchies that exist in real 3D indoor scenes~\cite{tang2023diffuscene}.

To address these limitations, we propose \model, a framework that formulates indoor scene synthesis as an order-aware sequential learning problem. \model organizes the inherently unordered collection of scene objects into structured, ordered hierarchical scene trees and forests. Specifically, \model first establishes orderings that prioritize the placement of dominant furniture pieces before the associated secondary objects, aligning with intuitive spatial reasoning principles for scene composition.

As illustrated in Figure~\ref{fig:teaser} (bottom left), \model used a clustering-based algorithm to parse the scene into a tree, which is then linearized into an ordered sequence via breadth-first traversal. To handle flexible objects like cabinets that can belong to multiple functional zones, as depicted in Figure~\ref{fig:teaser} (bottom right), we extend this to an ensemble of trees forming a scene forest representation. With these derived ordering, \model employs a transformer coupled with a denoising strategy. At inference time, \model generates plausible 3D scenes auto-regressively by sequentially placing furniture instances guided by the predicted order as shown in Figure~\ref{fig:teaser} (top).

We demonstrate the capability of \model to synthesize more realistic scenes using the 3D-FRONT dataset, outperforming the top-performing baseline with an average margin of FID score of $2.58$ and a KL score of $1.78$. Additionally, we illustrate how \model enhances scene rearrangement and completion tasks. Our extensive ablation studies further confirm that order-awareness as a prior significantly improves the generation of 3D indoor scenes.

\section{Related Works}

\noindent\textbf{Scene synthesis with handcrafted priors.} 
Early works {reasoned scene synthesis with various probabilistic models over scene exemplars in the view of object functionality criteria~\cite{ fisher2012example, zhao2016relationship, ma2018language} and human activities~\cite{fsl15, mlz16, sch16}}. 
{For example, Fisher \emph{et al.}~\cite{fisher2012example} investigated Bayesian network and Gaussian mixture model to model object co-occurrence. }
Make-it-Home~\cite{yu2011make} that pioneered in progressive synthesis, offered an interactive layout modeling tool by optimizing cost function that encoding spatial relation between furniture objects.
{Ma \emph{et al.}~\cite{mlz16} formulated an action graph with nodes represented as human actions}, guiding interior sythesis from human activity.
{In contrast, our approach is an end-to-end differentiable pipeline and free of externally introduced elements previously proposed for scene synthesis.}

\noindent\textbf{Scene synthesis via graph representation.} 
Modeling scenes as graphs~\cite{wang2019planit, li2019grains, zhou2019scenegraphnet, gao2023scenehgn, wei2023lego, tang2023diffuscene, henderson2017automatic} has been an intuitive approach for scene synthesis and been extensively studied recently. 
GRAINS~\cite{li2019grains} proposed a recursive auto-encoder network for scene synthesis, where novel scenes are generated hierarchically. 
SceneGraphNet~\cite{zhou2019scenegraphnet} utilized message-passing graph networks to model {long-range} relationships {among} objects. 
SceneHGN~\cite{gao2023scenehgn} defined a fine-grained hierarchy in room-object-part order, allowing multi-level scene editing. 
LEGO-Net~\cite{wei2023lego} and DiffuScene~\cite{tang2023diffuscene} represented scene as a fully-connected graph with {d}enoising {d}iffusion {p}robabilistic {m}odels (DDPMs).
CommonScenes~\cite{zhai2023commonscenes} {modeled scene jointly with layout and shape via} latent diffusion models.
{Our work represents scene as directed rooted forest and learns scene synthesis with transformers. }

\noindent\textbf{Scene synthesis using language modeling.} 
Recently, great success have been made in language modeling~\cite{vaswani2017attention, radford2019language} for content generation~\cite{openai2023gpt4, esser2021taming, chang2023muse, chang2022maskgit, rubenstein2023audiopalm, li2023starcoder}. 
SceneFormer~\cite{wang2021sceneformer} modeled each scene object property with individual transformer network, where object order is predefined by class frequency. 
ATISS~\cite{paschalidou2021atiss} represented object properties as span using a single transformer network and removed positional encoding for order permutation-invariance. 
CLIP-Layout~\cite{liu2023clip} learned to synthesize style-consistent indoor scenes with multi-modal CLIP~\cite{radford2021learning} encoder.
LayoutGPT~\cite{feng2023layoutgpt} leveraged {rich visual concepts and notable zero-shot capabilities} of large language models (LLMs), \emph{i.e.} ChatGPT~\cite{chatgpt}.
COFS~\cite{cofs} adapted masked language models and modeled scenes with a standard BART~\cite{lewis2020bart}-like generative model, formed by a bidirectional encoder over corrupted input and an auto-regressive decoder. 
In our method, we leverage the decoder-only casual transformer and denoising strategy to enhance the generation ability.

\noindent\textbf{Input as sequence or set.}
A significant limitation of language modeling is it can only be applied to problems whose inputs are represented as \emph{sequences}.
Hence, many research efforts~\cite{xu2018graph2seq, vinyals2015order, eriguchi-etal-2016-tree, iyer-etal-2016-summarizing} have been made to perform mappings from different data structure, like set, to sequences.
Set2Seq~\cite{vinyals2015order} proposed read, process, write block to process the input set and find the optimal orderings while training, the results of which show that order matters in various tasks.
Tree2Seq~\cite{currey-heafield-2018-unsupervised} added unsupervised hierarchical structure on the source sentence to Seq2Seq model for improving low-resource machine translation.
CODE-NN~\cite{iyer-etal-2016-summarizing} showcased the effectiveness of transforming structured code into sequence coupled with LSTM~\cite{hochreiter1997long} for summarization.
Another line of research initiatives have been focusing on processing the \emph{set}-input data.
{DeepSets}~\cite{deepset} {handled} set {data} by ensuring permutation-invariance and enabling pooling over sets.
Set Transformer~\cite{set-transformer-lee19d} featured as induced set attention block and attentional pooling module to aggregate set input attributes. 
Since the indoor scenes do not provide explicit ordering, following Set2Seq, our method ventures to search the optimal priori choice of the indoor scene ordering for sequential modeling.

\section{Method}

\begin{figure*}[!ht]
    \centering
    \includegraphics[width=1.0\linewidth]{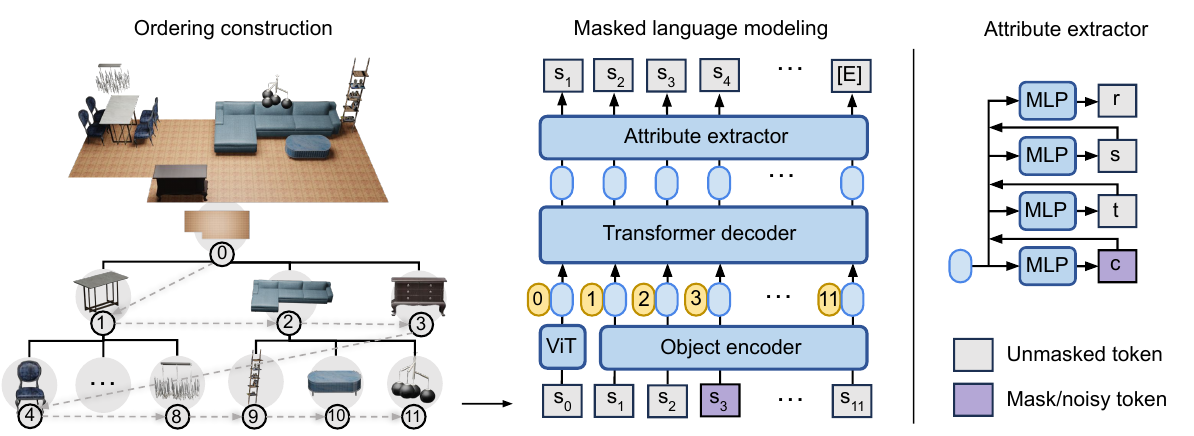}
    \caption{Training framework of our \model. On the left, we depict the construction of a tree/forest from parsing the scene and its subsequent flattening into a sequence through breadth-first search. The right panel illustrates our use of a causal transformer equipped with a denoising strategy for sequential data learning.}
    \label{fig:framework}    
\end{figure*}

Given a 2D floor/layout, 
we aim to develop a generative model to produce diverse and plausible object arrangements. 
Figure~\ref{fig:framework} shows the framework of our proposed method. 

\subsection{Ordering construction}

Scenes are represented as sets of oriented bounding boxes $\mathcal{O} = \{o_1, o_2, \ldots, o_n\}$, with each box $o_i = (c_i, t_i, b_i, r_i)$ containing a semantic class $c_i\in \mathbb{R}^C$, a 3D translation $t_i\in \mathbb{R}^3$, a bounding box size $b_i\in \mathbb{R}^3$, and a rotation angle $r_i\in \mathbb{R}$. 
To {enable transformer-based sequence-to-sequence learning,} we seek a permutation $\pi$ that transforms the set $\mathcal{O}$ into a sequence $\mathcal{S}=\pi(\mathcal{O})$.

\noindent\textbf{Scene tree.} 
In Figure~\ref{fig:framework} (left), the {implicit tree structure of the} living room is evident: 
Each subtree corresponds to a distinct functional zone, such as relaxation {zone} or dining {zone}, with the primary object forming the parent node and the ancillary objects as children.
This arrangement aligns {well} with intuitive spatial reasoning, where primary objects are positioned first, followed by their associated items. We posit this \emph{hierarchy}, which we refer to as scene tree ordering $\pi_T$, naturally suits indoor scene composition and aligns with practical spatial organization.

To realize the goal, we {build a tree with {Modified Euclidean Distance Clustering (MEDC)}, where on the {top-down} projected 2D bounding boxes $\overline{o}_i = (\overline{t}_i, \overline{s}_i)$}, we can calculate the distance between each {pair}, {formulating} a {distance} matrix ${M} = \{m_{ij}\}_{i,j=1}^N$ defined as:
\begin{equation}
    m_{ij} = d_{ij} + \lambda \cdot (1 - \text{GIoU}(\overline{ o}_i, \overline{ o}_j) ),
\end{equation}
where $\text{GIoU}(\cdot, \cdot) \in [-1, 1]$ is proposed in \cite{rezatofighi2019generalized} to evaluate of distance of {two} bounding boxes $\overline{ o}_i, \overline{ o}_j \in \mathbb R^4$, and $d_{ij}$ is the Euclidean distance between {two} bounding box centers. 
We employ the DBSCAN algorithm~\cite{dbscan} on the distance matrix to segment the furniture into multiple clusters and identify outliers.
Within each cluster, the largest {object} is designated as the root node, forming a subscene, with the remaining items as child nodes.
To eliminate inherent ordering among siblings, we randomly shuffle nodes under the same parent. Then, utilizing breadth-first search (BFS)~\cite{bundy1984breadth}, we linearize the shuffled tree into a sequence $\mathcal{S}_T = \pi_T(\mathcal{O})$.

\begin{figure*}
    \centering
    \includegraphics[width=\linewidth]{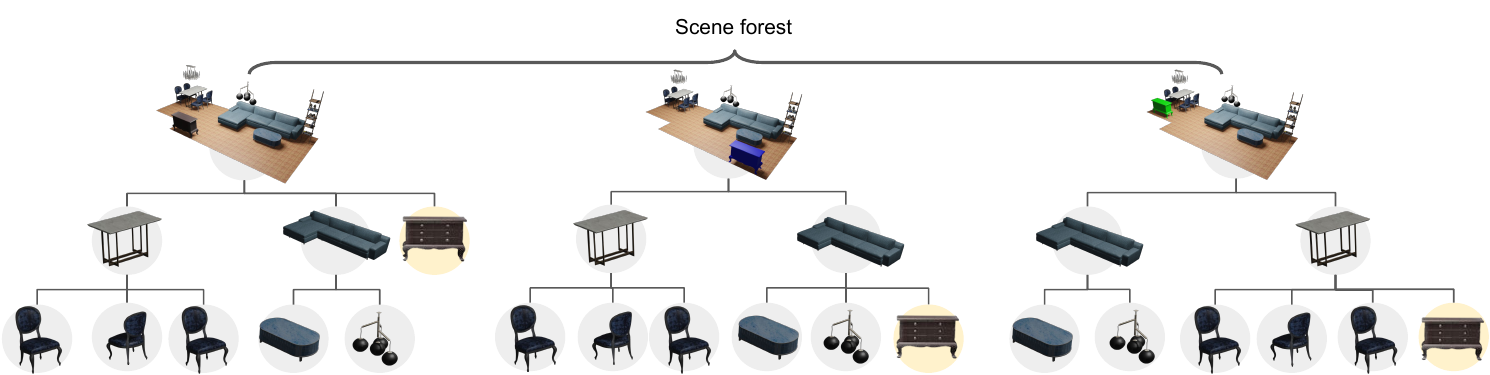}
    \caption{An example to illustrate the motivation of scene forest. The whole room is clearly divided to 2 subscenes according to the human activity. However, ``cabinet'' is an exception as it can reasonably belong to any subscene or the entire scene. Note that some items in the base tree are ignored for simplification.}
    \label{fig:forest}
\end{figure*}

\noindent\textbf{Scene forest.} 
As depicted in in Figure~\ref{fig:forest}, flexible furniture like cabinets capable of serving relaxation, dining, or the entire room, introduces ambiguity into the conventional scene tree representation. 
To resolve this, we propose an enriched tree structure that integrates these outliers by associating them with each potential parent node, resulting in s{an} ensemble of trees, as the forest scene representation.
During training, we randomly select a  tree from the forest (see Figure~\ref{fig:forest}) and employ BFS to convert it into a sequence, denoted as $\mathcal{S}_F = \pi_F(\mathcal{O})$. s{Note that this forest ordering approach, which produces inherently non-unique orderings, broadens the permutation space beyond what is possible with a single scene tree.}
{More details of the} algorithm are provided in the supplementary.

\subsection{Masked language modeling}
Upon establishing the scene priori ordering of the indoor scenes, see Figure~\ref{fig:framework} (right), our objective is to employ a transformer-based {generative model} to populate the ordered sequence $\mathcal{S} = \{s_i\}_{i=1}^N$. {The} framework primarily comprises four components, including layout encoder, object encoder, transformer decoder, and attribute extractor.

Two encoder {networks including layout encoder and object encoder} transform raw input into sequential embeddings. The layout encoder, a small vision transformer~\cite{dosovitskiy2020image, Gani2022HowTT}, {extracts} the binary layout mask $s_{0} \in \mathbb R^{64\times 64}$ into a start token ${x_0} \in \mathbb R^{512}$, establishing spatial constraints for object placement. The object encoder handles attributes $(c_i, {t}_i, {b}_i, r_i)$, {coupled with} sinusoidal positional encoding for {extracting} the discrete variable into a unified token ${x_i} = [\lambda(c_i); \psi(t_i); \psi(b_i); \psi(r_i)]\in \mathbb R ^{512}$.

The transformer decoder~\cite{radford2019language}, denoted as $f_\theta$, is tasked with the prediction of subsequent object embeddings. It achieves this by accumulating context from both the start layout token and {previous} object tokens, s{leverages} masked self-attention mechanisms. To enhance this process, we integrate absolute positional encodings to the sequence of object embeddings, thereby {enpowering} the model with crucial sequence order information:
\begin{equation}
\hat{{x}}_i = f_\theta({x}_{<i}; {x}_0),
\end{equation}
where $\hat{{x}}_i$ represents the predicted embedding of the next object, and ${x}_{<i}$ is the sequence of all previous object embeddings up to the $i$-th position.

Following previous work \cite{paschalidou2021atiss}, our feature extractor outputs a probability distribution for bounding box parameters, using $K$ logistic distributions for continuous parameters {including} position, size, and orientation:
\begin{equation}
p(h) = \sum_{j=1}^T\alpha_j \text{Logistic}(\mu_j, \sigma_j), 
\end{equation}
where $h$ is a component $(t_i, b_i, r_i)$, with $\alpha_j, \mu_j, \sigma_j$ {being} the logistic parameters. {And} discrete class probabilities are derived from logit vectors $l_c$ via the softmax function:
\begin{equation}
 p(c_i) = \text{Softmax}(l_c).
\end{equation}
Probability distributions are represented by $3K$-dim vectors for continuous, and $C$-dim for discrete class variables, where $C$ is the class {number}.

\noindent\textbf{Training with denoising.}
We employ input corruption techniques~\cite{Devlin2019BERTPO, gu2022vector, he2022masked} to mitigate overfitting in the transformer in both attribute- and object token-level. Specifically, a {predefined} percentage of the object embeddings are randomly substituted with a $\texttt{[MASK]}$ token. In addition, in the auto-regressive attribute prediction process, a {predefined} percentage of the ground truth categories are replaced with random categories, rather than consistent teacher forcing. 
We use the same 5$\%$ mask/noise rate for simplification.
Moreover, this design reduces the error propagation in sequential predictions.

\noindent\textbf{Inference.} During the inference phase, the process initiates with the layout embedding or start token. Subsequently, we employ an auto-regressive approach to iteratively sample attribute values from the distributions predicted for the subsequent object. Each newly generated object is concatenated with the preceding tokens, which then informs the subsequent generation step. This procedure is repeated until the end token is produced.

\noindent\textbf{{Object retrieval.}} Once a labeled oriented bounding box is sampled, we identify the matching furniture instance from the 3D-FUTURE dataset~\cite{fu20213dfuture} by selecting the closest size match within the predicted object category.

\subsection{Objective loss functions}
The language model is trained by minimizing the negative log-likelihood of the sequence joint distribution, factorized as the product of conditional probabilities across individual tokens~\cite{language2020bengio}:
\begin{equation} \label{eq1}
        \mathcal{L}_\theta = - \log \prod_{i=1}^N p_\theta(s_i| s_{<i}) 
        = - \sum_{i=1}^N \log  p_\theta(s_i| s_{<i}),
\end{equation}
where $p_\theta(s_i|s_{<i})$ is the cross entropy between the probability of the next object attributes predicted by model and the ground truth, given the previous $i$ tokens.

\subsection{Implementation details}
All experiments run on an NVIDIA RTX3090 GPU, with the AdamW~\cite{loshchilov2018decoupled} optimizer at a 1e-4 learning rate, without warm-up or decay strategies.
{For DBSCAN parameters, the eps and min-samples} are set to 0.15 {and} 2 respectively, with a GIoU weight of $\lambda$=0.02 applied to the distance matrix.
The attribute modeling employs a 10-component logistic mixture to accurately represent object distributions. 
All models are trained using a batch size of 128 across 1000 epochs, incorporating random rotations from 0$^{\circ}$ to 360$^{\circ}$ for {data} augmentation. We adopt standard practice for early stopping, evaluating against the validation metric every 10 epochs and selecting the best-performing iteration as the final model.
All layers are applied with a universal dropout rate of 0.1 to counter overfitting. 
{Details of the network architecture are provided in the supplementary.}

\section{Experiments}

\subsection{Experimental settings}
\noindent\textbf{Datasets.}
In alignment with prior work \cite{paschalidou2021atiss, cofs, tang2023diffuscene}, we utilize the 3D-FRONT dataset \cite{fu20213dfront} for training and evaluation, which includes around 10k professional 3D indoor scenes spanning bedrooms, libraries, living rooms, and dining rooms. Following {the preprocessing steps of} ATISS~\cite{paschalidou2021atiss}, the dataset is split into subsets with 3879/162 bedrooms, 230/56 libraries, 621/270 living rooms, and 723/177 dining rooms for training/validation. 
Given the limited library/living room/dining room data, we employ a pretrained bedroom model for initial weights for each room type to improve performance.

\noindent\textbf{Metrics.}
In line with established works~\cite{cofs, paschalidou2021atiss}, we employ various metrics for performance assessment, including Fréchet Inception Distance (FID)~\cite{heusel2017gans}, classification accuracy score (CAS), and categorical Kullback-Leibler {(KL)} divergence. The evaluation protocol involves rendering the indoor scenes into 256$\times$256 orthographic maps and calculating the CAS/FID scores against the ground truth.

\noindent\textbf{Baselines.}
Our method is benchmarked against recent advancements, including FastSynth~\cite{ritchie2019fast}, SceneFormer~\cite{wang2021sceneformer}, ATISS~\cite{paschalidou2021atiss}, COFS~\cite{cofs}, LayoutGPT~\cite{feng2023layoutgpt}, and DiffuScene~\cite{tang2023diffuscene}. Notably, DiffuScene utilizes top-down semantic maps for rendering 3D scenes, while LayoutGPT computes FID using images rendered from four distinct camera viewpoints. To ensure fair comparisons, we reproduce the experiment of these two methods using their official implementations. We do not report the results of DiffuScene on the library type, since the original paper did not conduct this experiment.

\subsection{Scene synthesis}

\noindent\textbf{Quantitative comparison.}
Our experiments on scene synthesis, when measured against baseline models, demonstrate the robustness of our approach. 
\begin{wraptable}{r}{4.4cm}
		\centering
		\resizebox{0.38\textwidth}{!}{%
			\begin{tabular}{l|cccc}
\hline
   Room type  & bedroom & living room & dining room & library \\
   \hline
  Num. of nodes &  5.00 & 11.7 & 10.9 & 4.61 \\
   Num. of trees & 1.27 & 2.83 & 2.51 & 1.14 \\
 \hline
\end{tabular}

		}
  \caption{Forest statistics.}
  \label{tab:stat}
\end{wraptable}
As {described} in Table \ref{tab:uncond}, \model surpasses all baselines with superior KL divergence scores, implying a closer match to the ground truth. Furthermore, \model achieves the most favorable scores in both the FID and CAS, indicating its ability to render more {realistic} scenes. 
Table~\ref{tab:stat} shows the basic statistics of the scene forest: living room and dining room contains the most nodes and trees, which explains performance improvement more substantial in the two room types.
In terms of computational efficiency, as illustrated in Table~\ref{tab:eff}, our model demonstrates competitive inference times and requires the fewest parameters (48$\%$ fewer than the model with the second fewest parameters) among all methods compared. It is noteworthy {that we} employ a compact transformer decoder and a small ViT-based layout encoder (2.58M) to mitigate overfitting and {improve} efficiency. 
\setlength{\tabcolsep}{4pt}
\begin{table*}[!hbt]
	\renewcommand\arraystretch{1.2}
	\begin{center}
  \resizebox{\linewidth}{!}
    {
		\begin{tabular}{l|ccc |ccc|ccc|ccc}
                \hline
			  \multirow{2}{*}{Method} & \multicolumn{3}{c|}{\cellcolor{white!20}{Bedroom}}   & \multicolumn{3}{c|}{\cellcolor{white!20}{Dining room}} & \multicolumn{3}{c|}{\cellcolor{white!20}{Living room}}  & \multicolumn{3}{c}{\cellcolor{white!20}{Library}}\\
			 & KL & FID  & CAS ($\%$)    & KL& FID  & CAS ($\%$)  
             & KL & FID  & CAS  ($\%$)
             & KL & FID  & CAS  ($\%$)  \\ 
			\hline 

            FastSyn~\cite{ritchie2019fast}
                  & 6.4 & 88.1  & 88.3  
                  &  51.8 & 58.9  & 93.5 
                  & 17.6 & 66.6  & 94.5  
                  & 43.1 & 86.6 & 81.5\\
            SceneFormer~\cite{wang2021sceneformer}
                  & 5.2 & 90.6 & 97.2  
                  &  36.8 & 60.1 & 71.3 
                  & 31.3 & 68.1 & 72.6   
                  & 23.2 & 89.1 & 88.0\\
            LayoutGPT \cite{feng2023layoutgpt}
                  & 17.5 & 68.1 & 60.6 %
                 & {---}  & {---} & {---} 
                 & 14.0 & 76.3  & 94.5  %
                 & {---} & {---}  & {---}  \\
             
            ATISS~\cite{paschalidou2021atiss}
                   & 8.6 & 73.0  & 61.1 
                  &  15.6 & 47.6  & {69.1} 
                  & 14.1 & 43.3  & 76.4  
                  & 10.1 & {75.3} &  {61.7} \\

            COFS~\cite{cofs}
                  & {5.0} & 73.2  & 61.0 
                  &  9.3 & {43.1}   & 76.1
                  & {8.1}  & {35.9}   & 78.9 
                  & {6.7} & 75.7 & 66.2\\

            DiffuScene \cite{tang2023diffuscene}
                 & {5.1} & {69.0} & {59.7}   
                 & {7.9} & 45.8   & {70.6}
                 & 8.3 & 38.2  & {75.1} 
                 & --- & ---  &  --- \\

            Ours
                  & \textbf{4.2} & \textbf{67.9}& \textbf{58.3 }
                  & \textbf{5.5}& \textbf{40.2}& \textbf{65.6}
                  & \textbf{5.9}& \textbf{35.2}& \textbf{68.0}
                  & \textbf{5.2}& \textbf{69.1}& \textbf{57.3}\\
        \hline
        \end{tabular}
        }
        \caption{Quantitative comparison with the state-of-the-art methods~\cite{paschalidou2021atiss, cofs, tang2023diffuscene, wang2021sceneformer,ritchie2019fast} on the task of scene synthesis. Note that for FID and KL, lower is better, and for CAS, the score closer to $50\%$ is better.}
        \label{tab:uncond}
        \end{center}
\end{table*}

\begin{figure*}[!ht]
    \centering
    \includegraphics[width=\linewidth]{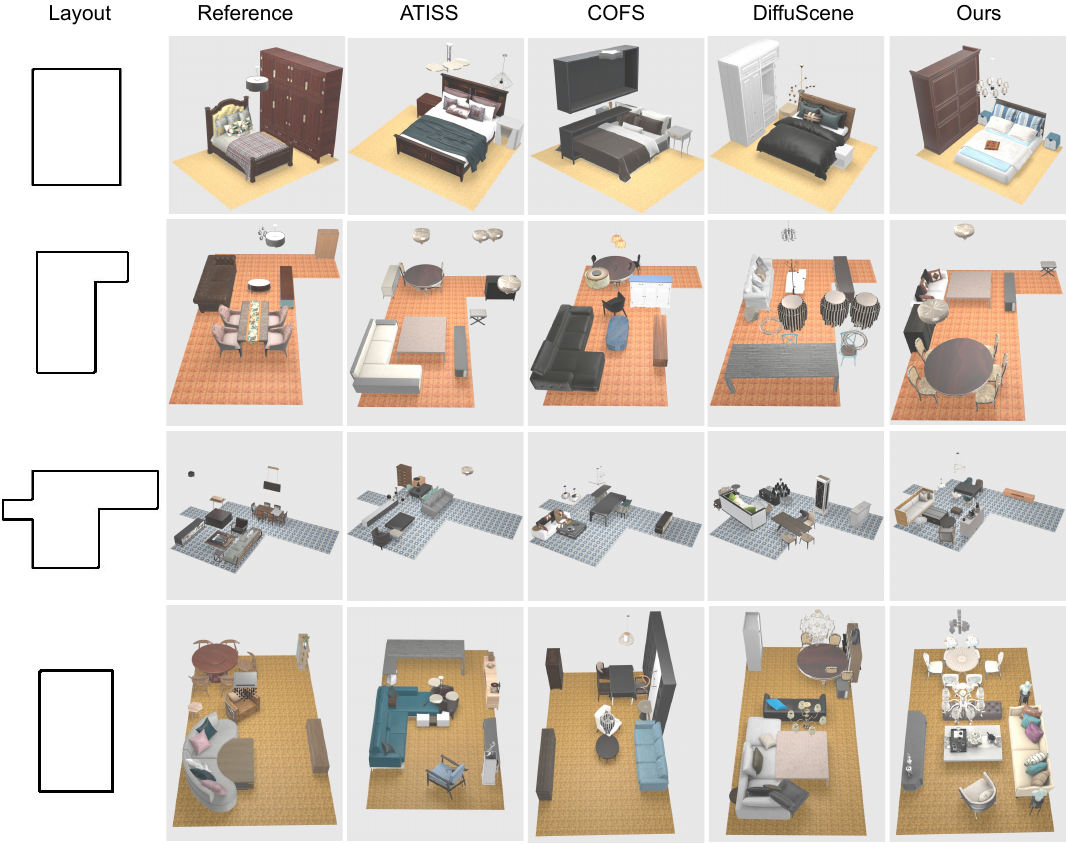}
    \caption{Qualitative comparison with the-state-of-the-art methods~\cite{paschalidou2021atiss, cofs, tang2023diffuscene} on scene synthesis for three type of scenes: bedrooms (1st row), living room (2nd and 3rd rows) and dining room (4th row). Note that reference is the scene from dataset with the same floor plan.
}
    \label{fig:comparison}
    \vspace{-3mm}
\end{figure*}
\noindent\textbf{Qualitative comparison.}
In Figure~\ref{fig:comparison}, we illustrate the {visual results} of scene synthesis. To ensure a balanced evaluation, the same randomly sampled room layout serves as the conditional input for DiffuScene, ATISS, COFS, and our \model across various room types. The comparison reveals that scenes synthesized by \model exhibit greater alignment with the given floor plans and demonstrate a reduced tendency to place furniture beyond room boundaries.

\begin{table}[!bt]
	\centering
	\begin{subfigure}{0.5\linewidth}
		\centering
		\resizebox{\textwidth}{!}{%
			\begin{tabular}{l|cccc}
\hline
   Method & Bedroom  & Living room & Dining room & Library \\
   \hline 
ATISS  & 20.2 & 10.1 & 13.9 & 26.5 \\
COFS  & 13.4 & 21.2 & 8.5 & 25.0\\
DiffuScene  & 23.1 & 16.7 & 19.6 & --- \\
Ours  & \textbf{43.3} & \textbf{52.0} & \textbf{58.0}& \textbf{48.5} \\
 \hline
\end{tabular}

		}\caption{User study} 
		\label{tab:user}
	\end{subfigure}
	\begin{subfigure}{0.49\linewidth}
		\centering
		\resizebox{\textwidth}{!}{%
			\begin{tabular}{l|cccc}
\hline
   Method   &  ATISS & COFS  & DiffuScene & Ours\\
   \hline 
   Parameters (MB)     & 36.1 & 19.4 & 74.1 & \textbf{9.99} \\
   Inference rate (s)  & 0.204& \textbf{0.129} & 34.9 & {0.160}\\
 \hline
\end{tabular}

		}
		\vspace{0.27cm}
		\caption{Efficiency comparisons}
		\label{tab:eff}
	\end{subfigure}\\
	\vspace{-0.2cm}
	\caption{
 \textbf{Additional quantitative comparison with the state-of-the-arts.}
    }
\end{table}

\noindent\textbf{Perceptual study.}
{In Table~\ref{tab:user}, we conduct user study by 37 participants who assessed the realism of scenes generated by DiffuScene, ATISS, and COFS across 51 randomly selected rooms of each type for evaluation. 
The perceptual analysis reveals that more than 50$\%$ of our generated scenes for living rooms and dining rooms are considered realistic than others. Furthermore, our results gain a predominant preference in bedrooms and libraries.}

\noindent\textbf{Analysis.}
{ATISS utilizes a transformer encoder without positional encoding and randomizes object order while training to achieve approximate} permutation invariance. 
Similarly, COFS posits that the layout is inherently unordered, leveraging BART, a masked language model, to underpin this assumption. DiffuScene represents scenes as fully-connected graphs and learns to denoise over Gaussian noise.
However, these baselines do not account for the inherent ordering among objects. This oversight often results in predictions that place objects too close to one another, leading to frequent intersections and {thereby affecting} scene realism. 
On the other hand, our {method} introduces multiple {strategies} to mitigate overfitting: 1) {our} scene forest offers a priori ordering that guides the generative process; 2) we adopt a decoder-only architecture, which not only reduces the {network capacity} but also exhibits enhanced generative performance over encoder-decoder frameworks~\cite{fu2023decoder} such as BART; 3) the denoising training strategy improves the generalization {ability of the model}.

\subsection{Discussion on the order prior}
In Table~\ref{tab:abl}, we incorporate different orderings scheme with the auto-regressive generating network, in which each scene can be represented either with a single (the first three columns) or multiple (the last three columns) ordered sequences. 
We also report two statistics for the set of sequences: diversity means that the average number of ordered sequences; inconsistency is evaluated by the average hamming distance between two sequence pair within the set. 

\begin{table}[]
    \centering
    \resizebox{\linewidth}{!}
    {
    \begin{tabular}{l|cccccc}
    \hline
       Order (set)   & Random(single) & Fixed & Tree+BFS & Random(multiple) & Forest+DFS & Forest+BFS \\
       \hline
         Diversity  & 1 & 1 & 1 & $\infty$  &  2.83  & 2.83 \\
          Inconsistency  &  0 & 0 & 0 & 9.54 & 4.01 &  1.87 \\
          \hline
       KL $\downarrow$  &  20.0 &  17.9 & \underline{7.90} & 13.1 & 9.40 &  \textbf{5.90} \\
      FID $\downarrow$  &  49.4 & 49.8  & \underline{36.1} & 43.3 &  40.5 & \textbf{35.2} \\
       CAS ($\%$)  & 83.7 & 80.1 & \underline{68.1} & 76.4 & 71.7 & \textbf{68.0}\\
     \hline
    \end{tabular}
    }
    \caption{{Ablation study:} 
    comparison of KL, FID and CAS of different order settings in living room scenario. 
    The second-best score is underscored.
    }
    \label{tab:abl}
\end{table}

\begin{figure*}
    \centering
    \includegraphics[width=\linewidth]{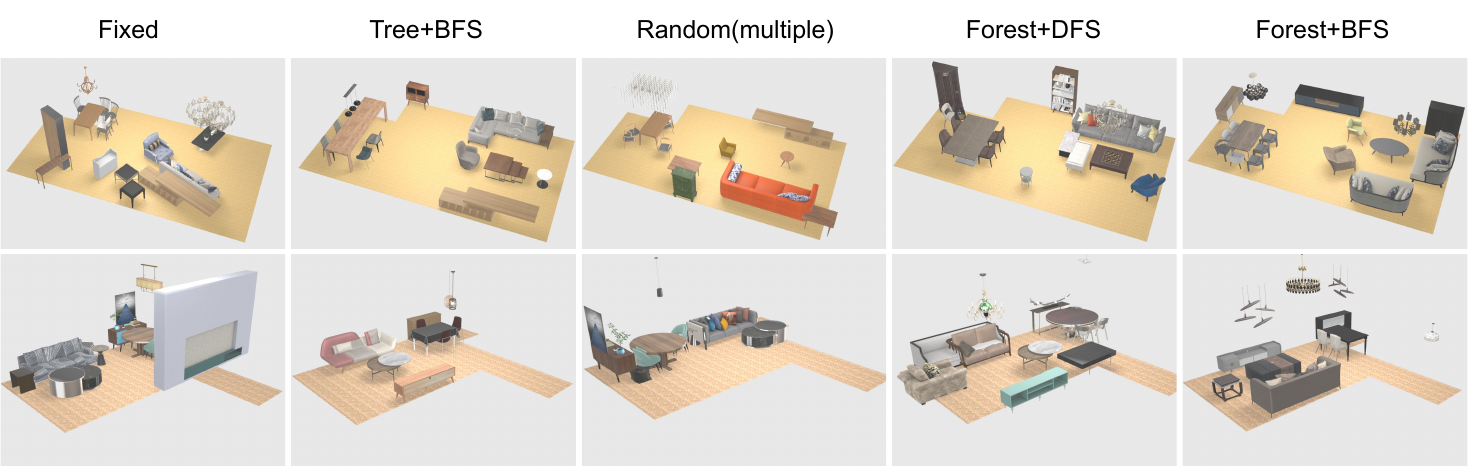}
    \caption{{Ablation study:} visual comparison of different orderings. While random order and fixed order can not provide appropriate prior, our tree-guided order benefits the scene synthesis, generating plausible scenes. The forest representation further enhances the scene diversity and realism. }
    \label{fig:ablation}
\end{figure*}

\noindent\textbf{Superiority of forest ordering.}
We observe that random permutation and fixed order~\cite{wang2021sceneformer} using frequency-based arrangement perform worse than tree-guided order. This indicates that an optimal ordering could be beneficial for scene modeling.
The further improvement in KL from the tree to forest demonstrates the benefit of our introducing scene forest, which explicitly model the flexible objects.
When comparing breadth-first (BFS) and depth-first (DFS) traversal methods for scene tree sequences, we find that BFS yields superior results due to the greater consistency within sets of BFS-derived sequences.
Additionally, we examine the ordering strategy that is utilized in ATISS~\cite{paschalidou2021atiss} that shuffles the sequence before fed into the network. This approach, representing a scene as multiple random sequences, shows improvement over the single random case, attributed to data augmentation and the resulting high diversity. However, its performance is constrained by sequence inconsistency (9.54) within the set.
Supported by qualitative evidence in Figure~\ref{fig:ablation}, these findings underscore the effectiveness of our proposed forest ordering in generating more realistic and diverse object arrangements. Notably, the placement of primary furniture is less accurate in the DFS scenario compared to the BFS scenario.

\noindent\textbf{On the methods for forest formation/order construction.}
The goal of forest formation is to reconstruct the scene hierarchy from the given object representations. Several studies~\cite{tang2019learning, zellers2018neural, socher2011parsing} have explored learning tree structures or scene graphs from natural images. Consequently, we evaluate the scene parsing efficacy of pure Euclidean Distance Clustering (EDC), Modified Euclidean Distance Clustering (MEDC), and a learning-based baseline, VCTree~\cite{tang2019learning} that was originally designed for parsing natural images. We assess reconstruction accuracy using Average Hierarchical Distance (AHD), which calculates similarity by comparing sets formed at each tree depth between the reconstructed and human-annotated trees. AHD averages these set-based similarity scores across all depths for a concise measure of accuracy. As indicated in Table~\ref{tab:parsing}, while the proposed MEDC method is straightforward, it effectively captures underlying semantic relationships and offers computational efficiency advantages.

\begin{wrapfigure}{r}{6cm}
		\centering
		\resizebox{0.51\textwidth}{!}{%
			\includegraphics[width=\linewidth]{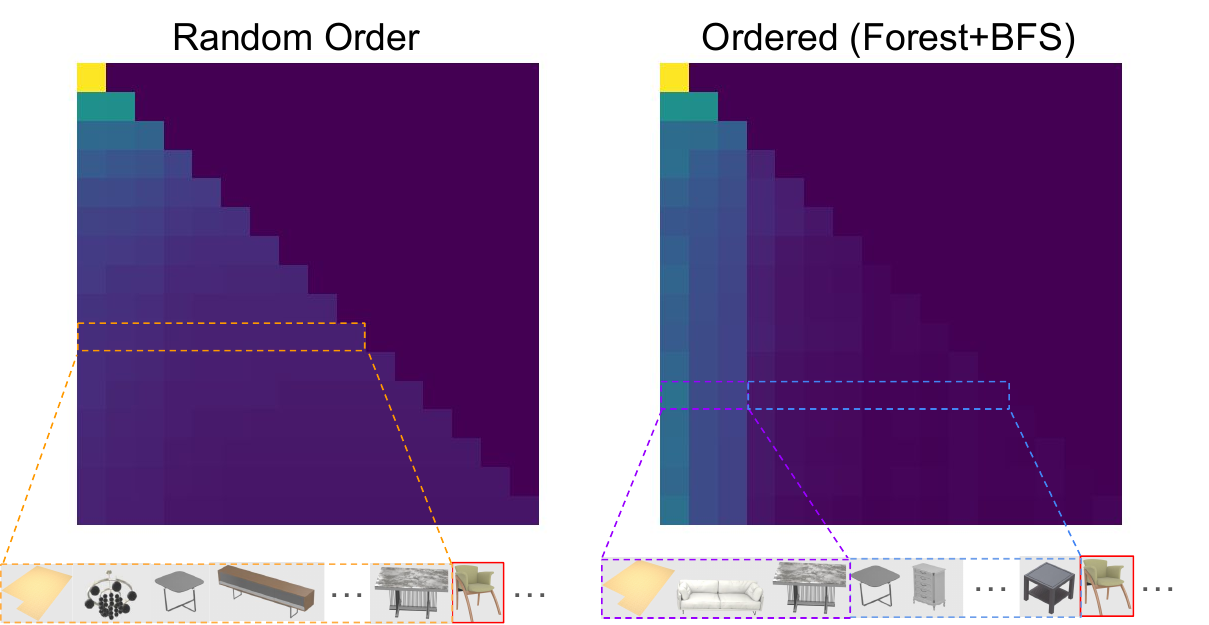}
		}
  \caption{Attention map for different ordering. The dash line boxes indicates the previous objects that are used to predict the next objects that are annotated with red solid boxes.}
  \label{fig:attn}
  \vspace{-5mm}
\end{wrapfigure}

\noindent\textbf{Attention visualization.}
Figure~\ref{fig:attn} presents a visualization of the attention heatmaps trained under multiple random ordering and scene forest ordering. It is evident that with our scene forest ordering, pivotal objects receive heightened attention scores in the next object token prediction. This inherent bias of underlying model confirms that the order prior affects the model, also explains why the object embeddings are predicted more accurately. Conversely, in the unordered scenario, the distribution of attention scores is notably uniform across the object sequence {without attention efficacy}.

\subsection{More ablation study}
\noindent\textbf{On the positional encoding.}
A core component of Transformers is the {positional encoding} mechanism that represents the order of input sequence~\cite{ke2020rethinking}. We compare the results of transformer decoder variant without positional encoding and those incorporating relative~\cite{shaw2018self} and absolute~\cite{vaswani2017attention} positional encoding. In Table~\ref{tab:pe}, the results clearly demonstrate that incorporating positional encoding facilitates learning of order, with the model variant employing absolute positional encoding achieving marginally better generation outcomes.

\noindent\textbf{On denoising strategy.}
Table~\ref{tab:de} shows the effectiveness of our denoising strategy. By introducing a small rate (0.05 or 0.1) of masked and random tokens, we effectively prevent overfitting and enhance model generalization. Conversely, at a higher mask/noise rate (0.2), we find that achieving convergence of training loss becomes challenging. 
\begin{table}[!bt]
	\centering
 	\begin{subfigure}[b]{0.28\linewidth}
		\centering
		\resizebox{\textwidth}{!}{%
			\begin{tabular}{l|cc}
\hline
    Methods   & AHD & Inference time(ms) \\
   \hline 
   MEDC  & \textbf{0.84} & 0.827\\
   EDC   &  0.53 & 0.777 \\
    VCTree~\cite{tang2019learning}   & 0.77 & 56.1 \\
 \hline
\end{tabular}

		}
        \caption{On different scene parsing methods.} 
		\label{tab:parsing}
	\end{subfigure}
    \hfill
	\begin{subfigure}[b]{0.34\linewidth}
		\centering
		\resizebox{\textwidth}{!}{%
			\begin{tabular}{l|ccc}
\hline
   Positional Encoding type   & KL & FID & CAS ($\%$) \\
   \hline 
   w/o  & 11.1 & 41.2 & 74.3\\
   w. absolute~\cite{vaswani2017attention}     & \textbf{5.9} & \textbf{35.2} & \textbf{68.0}\\
   w. relative~\cite{shaw2018self}    & 6.2 & 36.7  & 70.1 \\
 \hline
\end{tabular}

		}
        \caption{On the type of positional encoding.} 
		\label{tab:pe}
	\end{subfigure}
     \hfill
	\begin{subfigure}[b]{0.34\linewidth}
		\centering
		\resizebox{\textwidth}{!}{%
			\begin{tabular}{l|ccccc}
\hline
   Mask/noise rate & 0 & 0.05 & 0.1 & 0.15 & 0.2  \\
   \hline
   KL $\downarrow$ & 6.4 & \textbf{5.9} & 6.1 & 8.8 &  12.2 \\
   FID $\downarrow$& 35.6 & 35.2  & \textbf{34.4} &  39.6 & 42.5 \\
   CAS ($\%$) & 69.2 & 68.0 & \textbf{66.7} & 73.1 & 74.5\\
 \hline
\end{tabular}

		}
		\caption{On the mask/noise rate in denoising training scheme.}
		\label{tab:de}
	\end{subfigure}\\
	\vspace{-0.2cm}
	\caption{\textbf{Ablation study}.}
        \vspace{-0.8cm}
\end{table}

\begin{figure}
    \centering
    \includegraphics[width=0.95\linewidth]{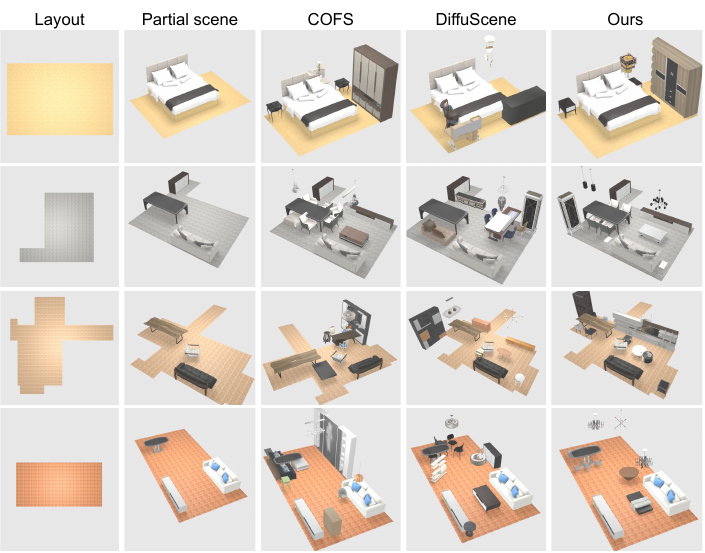}
    \caption{\textbf{Scene completion --} Comparison with COFS and DiffuScene for bedroom (1st row) and living room (the bottom three rows).}
    \label{fig:completion}
    \vspace{-4mm}
\end{figure}

\begin{figure}
    \centering
    \includegraphics[width=\linewidth]{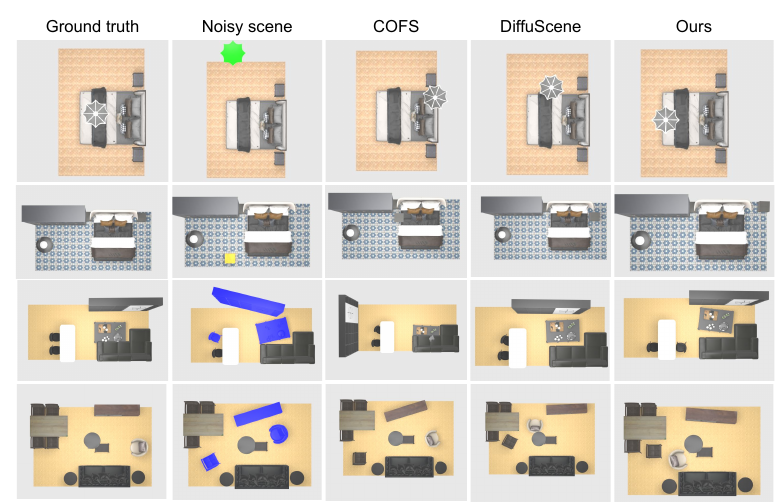}
    \caption{\textbf{Scene rearrangement --} Comparison with COFS and DiffuScene for living room (the top two rows) and dining room (the bottom two rows).}
    \label{fig:rearange}
    \vspace{-3mm}
\end{figure}

\subsection{Application in downstream tasks}
With the sequential generation framework, our method is easily applicable to various downstream tasks, such as scene completion and {rearrangement}. 

\noindent\textbf{Scene completion.}
 We retain the ground truth of the first $N$ objects, predicting subsequent tokens auto-regressively util the sequence concludes.
In Figure~\ref{fig:completion}, we compare against the state-of-the-art method, DiffuScene~\cite{tang2023diffuscene} and COFS~\cite{cofs}, on the task. Unlike DiffuScene and COFS, which omits essential items like lights (1st row) and introduces misplaced elements such as dining tables and cabinets (2nd row), our approach (3nd {column}) yields clean and coherent scenes due to the awareness of the key furniture placement.

\noindent\textbf{Rearangement.}  
Our method corrects one or multiple failure cases by resampling {the position} of an object considering prior inputs. Figure~\ref{fig:rearange} illustrates our success in adjusting the location of a night stand (2nd row) and optimally placing the bookshelf and chairs (the last row), an improvement over the inability of DiffuScene to correct these placements.
\section{Conclusion, Limitations and Future Work}
\begin{figure}
    \centering
    \includegraphics[width=\linewidth]{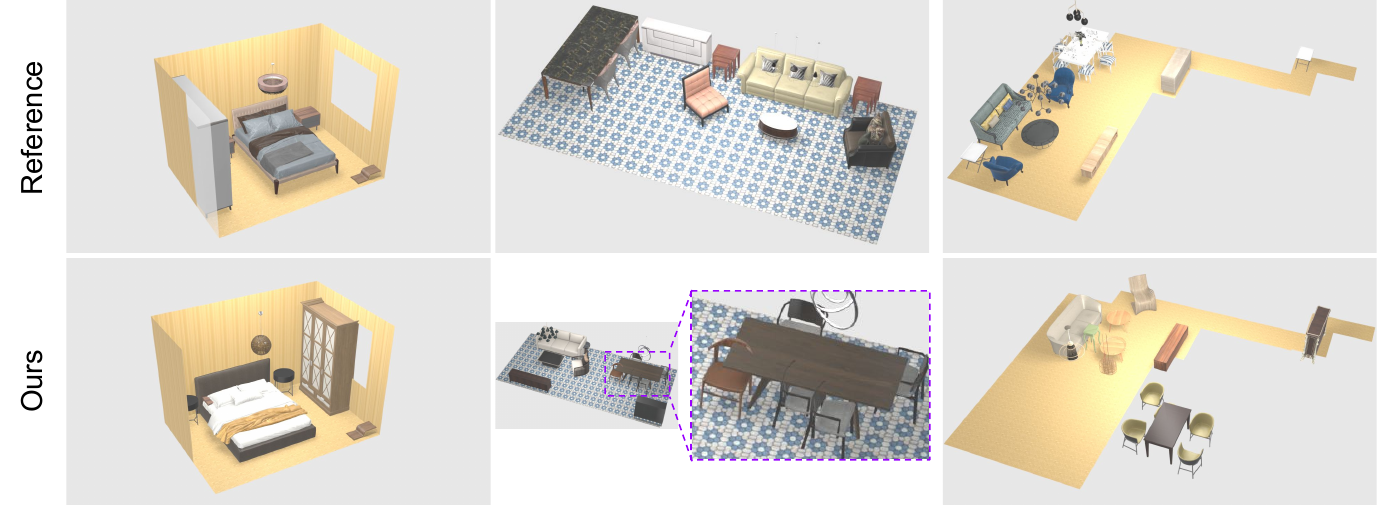}
    \caption{\textbf{Failure cases --} Neglecting window placement (left); overlapping furniture arrangements (mid); objects placed out of boundary in non-standard layouts (right).}
    \label{fig:failure}
    \vspace{-3mm}
\end{figure}

We have introduced \model, a novel framework to synthesize indoor scenes via sequential modeling. 
In contrast with previous works that neglect ordering, we leverage a parsing tree/forest and breadth-first search (BFS) to explore the implicit scene ordering, train the underlying network in an order-aware manner, and validate that \emph{order matters} in scene modeling through extensive experiments including scene synthesis, completion and rearrangement.

Figure~\ref{fig:failure} highlights three primary limitations of our method.
The left column indicates that the cabinet is blocking the window, creating undesirable scenes. This occurs since our model currently not factoring doors and windows as additional condition.
In the mid column, we observe occasional instance of object overlaps due to a lack of spatial constraints on relative positioning.
As demonstrated by the right, the model struggles with generating plausible results for highly complex layouts, partially due to the limited diversity of training data. 

In addition to addressing the difficulties suggested by the failure cases, the future work should explore towards the optimal order for sequential generation. For example, this could involve integrating learning-to-ordering module into an end-to-end network by joint optimizing the ordering and the likelihood objective.

\noindent\textbf{Acknowledgements:} We thank all anonymous reviewers and area chairs for their valuable comments, Jing Nathan Yan and Prof. Jing Liao for insightful discussions.


\bibliographystyle{splncs04}
\bibliography{egbib}

\begin{thebibliography}{10}
\providecommand{\url}[1]{\texttt{#1}}
\providecommand{\urlprefix}{URL }
\providecommand{\doi}[1]{https://doi.org/#1}

\bibitem{ikea}
\url{https://www.ikea.com/pt/en/customer-service/services/planning-consultation}

\bibitem{language2020bengio}
Bengio, Y., Ducharme, R., Vincent, P.: A neural probabilistic language model. In: NeurIPS (2000)

\bibitem{bundy1984breadth}
Bundy, A., Wallen, L.: Breadth-first search. Catalogue of Artificial Intelligence Tools  (1984)

\bibitem{Matterport3D}
Chang, A., Dai, A., Funkhouser, T., Halber, M., Niessner, M., Savva, M., Song, S., Zeng, A., Zhang, Y.: Matterport3{D}: Learning from {RGB-D} data in indoor environments. 3DV  (2017)

\bibitem{chang2023muse}
Chang, H., Zhang, H., Barber, J., Maschinot, A., Lezama, J., Jiang, L., Yang, M.H., Murphy, K., Freeman, W.T., Rubinstein, M., et~al.: Muse: Text-to-image generation via masked generative transformers. arXiv preprint arXiv:2301.00704  (2023)

\bibitem{chang2022maskgit}
Chang, H., Zhang, H., Jiang, L., Liu, C., Freeman, W.T.: Mask{GiT}: Masked generative image transformer. In: CVPR (2022)

\bibitem{Siddhartha2013uist}
Chaudhuri, S., Kalogerakis, E., Giguere, S., Funkhouser, T.: Attribit: content creation with semantic attributes. In: UIST (2013)

\bibitem{currey-heafield-2018-unsupervised}
Currey, A., Heafield, K.: Unsupervised source hierarchies for low-resource neural machine translation. In: ACL (2018)

\bibitem{procthor}
Deitke, M., VanderBilt, E., Herrasti, A., Weihs, L., Salvador, J., Ehsani, K., Han, W., Kolve, E., Farhadi, A., Kembhavi, A., Mottaghi, R.: {ProcTHOR: Large-Scale Embodied AI Using Procedural Generation}. In: NeurIPS (2022)

\bibitem{devaranjan2020meta}
Devaranjan, J., Kar, A., Fidler, S.: Meta-sim2: Unsupervised learning of scene structure for synthetic data generation. In: ECCV (2020)

\bibitem{Devlin2019BERTPO}
Devlin, J., Chang, M.W., Lee, K., Toutanova, K.: {BERT}: Pre-training of deep bidirectional transformers for language understanding. In: NAACL (2019)

\bibitem{dosovitskiy2020image}
Dosovitskiy, A., Beyer, L., Kolesnikov, A., Weissenborn, D., Zhai, X., Unterthiner, T., Dehghani, M., Minderer, M., Heigold, G., Gelly, S., et~al.: An image is worth 16x16 words: Transformers for image recognition at scale. In: ICLR (2021)

\bibitem{eriguchi-etal-2016-tree}
Eriguchi, A., Hashimoto, K., Tsuruoka, Y.: Tree-to-sequence attentional neural machine translation. In: ACL (2016)

\bibitem{esser2021taming}
Esser, P., Rombach, R., Ommer, B.: Taming transformers for high-resolution image synthesis. In: CVPR (2021)

\bibitem{dbscan}
Ester, M., Kriegel, H.P., Sander, J., Xu, X., et~al.: A density-based algorithm for discovering clusters in large spatial databases with noise. In: KDD (1996)

\bibitem{feng2023layoutgpt}
Feng, W., Zhu, W., Fu, T.j., Jampani, V., Akula, A., He, X., Basu, S., Wang, X.E., Wang, W.Y.: Layout{GPT}: Compositional visual planning and generation with large language models. arXiv preprint arXiv:2305.15393  (2023)

\bibitem{fisher2012example}
Fisher, M., Ritchie, D., Savva, M., Funkhouser, T., Hanrahan, P.: Example-based synthesis of {3D} object arrangements. ACM TOG  (2012)

\bibitem{fsl15}
Fisher, M., Savva, M., Li, Y., Hanrahan, P., Nie{\ss}ner, M.: Activity-centric scene synthesis for functional {3D} scene modeling. ACM TOG  (2015)

\bibitem{fu20213dfront}
Fu, H., Cai, B., Gao, L., Zhang, L.X., Wang, J., Li, C., Zeng, Q., Sun, C., Jia, R., Zhao, B., et~al.: {3D-FRONT}: 3{D} furnished rooms with layouts and semantics. In: CVPR (2021)

\bibitem{fu20213dfuture}
Fu, H., Jia, R., Gao, L., Gong, M., Zhao, B., Maybank, S., Tao, D.: {3D-FUTURE}: 3{D} furniture shape with texture. IJCV  (2021)

\bibitem{fu2023decoder}
Fu, Z., Lam, W., Yu, Q., So, A.M.C., Hu, S., Liu, Z., Collier, N.: Decoder-only or encoder-decoder? {I}nterpreting language model as a regularized encoder-decoder. arXiv preprint arXiv:2304.04052  (2023)

\bibitem{Gani2022HowTT}
Gani, H., Naseer, M., Yaqub, M.: How to train vision transformer on small-scale datasets? In: BMVC (2022)

\bibitem{gao2023scenehgn}
Gao, L., Sun, J.M., Mo, K., Lai, Y.K., Guibas, L.J., Yang, J.: {SceneHGN}: Hierarchical graph networks for {3D} indoor scene generation with fine-grained geometry. IEEE TPAMI  (2023)

\bibitem{gu2022vector}
Gu, S., Chen, D., Bao, J., Wen, F., Zhang, B., Chen, D., Yuan, L., Guo, B.: Vector quantized diffusion model for text-to-image synthesis. In: CVPR (2022)

\bibitem{he2022masked}
He, K., Chen, X., Xie, S., Li, Y., Doll{\'a}r, P., Girshick, R.: Masked autoencoders are scalable vision learners. In: CVPR (2022)

\bibitem{he2016deep}
He, K., Zhang, X., Ren, S., Sun, J.: Deep residual learning for image recognition. In: CVPR (2016)

\bibitem{henderson2017automatic}
Henderson, P., Subr, K., Ferrari, V.: Automatic generation of constrained furniture layouts. arXiv preprint arXiv:1711.10939  (2017)

\bibitem{heusel2017gans}
Heusel, M., Ramsauer, H., Unterthiner, T., Nessler, B., Hochreiter, S.: {GAN}s trained by a two time-scale update rule converge to a local nash equilibrium. NeurIPS  (2017)

\bibitem{hochreiter1997long}
Hochreiter, S., Schmidhuber, J.: Long short-term memory. Neural Computation  (1997)

\bibitem{iyer-etal-2016-summarizing}
Iyer, S., Konstas, I., Cheung, A., Zettlemoyer, L.: Summarizing source code using a neural attention model. In: ACL (2016)

\bibitem{jls12}
Jiang, Y., Lim, M., Saxena, A.: Learning object arrangements in {3D} scenes using human context. arXiv preprint arXiv:1206.6462  (2012)

\bibitem{kar2019meta}
Kar, A., Prakash, A., Liu, M.Y., Cameracci, E., Yuan, J., Rusiniak, M., Acuna, D., Torralba, A., Fidler, S.: Meta-sim: Learning to generate synthetic datasets. In: ICCV (2019)

\bibitem{ke2020rethinking}
Ke, G., He, D., Liu, T.Y.: Rethinking positional encoding in language pre-training. arXiv preprint arXiv:2006.15595  (2020)

\bibitem{koh2023simple}
Koh, J.Y., Agrawal, H., Batra, D., Tucker, R., Waters, A., Lee, H., Yang, Y., Baldridge, J., Anderson, P.: Simple and effective synthesis of indoor 3d scenes. In: AAAI (2023)

\bibitem{set-transformer-lee19d}
Lee, J., Lee, Y., Kim, J., Kosiorek, A., Choi, S., Teh, Y.W.: Set transformer: A framework for attention-based permutation-invariant neural networks. In: ICML (2019)

\bibitem{Leimer10.1145/3550469.3555425}
Leimer, K., Guerrero, P., Weiss, T., Musialski, P.: Layoutenhancer: Generating good indoor layouts from imperfect data. In: SIGGRAPH Asia (2022)

\bibitem{lewis2020bart}
Lewis, M., Liu, Y., Goyal, N., Ghazvininejad, M., Mohamed, A., Levy, O., Stoyanov, V., Zettlemoyer, L.: {BART}: Denoising sequence-to-sequence pre-training for natural language generation, translation, and comprehension. In: ACL (2020)

\bibitem{li2019grains}
Li, M., Patil, A.G., Xu, K., Chaudhuri, S., Khan, O., Shamir, A., Tu, C., Chen, B., Cohen-Or, D., Zhang, H.: {GRAINS}: Generative recursive autoencoders for indoor scenes. ACM TOG  (2019)

\bibitem{li2023starcoder}
Li, R., Allal, L.B., Zi, Y., Muennighoff, N., Kocetkov, D., Mou, C., Marone, M., Akiki, C., Li, J., Chim, J., et~al.: {StarCoder}: May the source be with you! arXiv preprint arXiv:2305.06161  (2023)

\bibitem{liu2023clip}
Liu, J., et~al.: {CLIP-Layout}: Style-consistent indoor scene synthesis with semantic furniture embedding. arXiv preprint arXiv:2303.03565  (2023)

\bibitem{loshchilov2018decoupled}
Loshchilov, I., Hutter, F.: Decoupled weight decay regularization. In: ICLR (2019)

\bibitem{mlz16}
Ma, R., Li, H., Zou, C., Liao, Z., Tong, X., Zhang, H.: Action-driven {3D} indoor scene evolution. ACM TOG  (2016)

\bibitem{ma2018language}
Ma, R., Patil, A.G., Fisher, M., Li, M., Pirk, S., Hua, B.S., Yeung, S.K., Tong, X., Guibas, L., Zhang, H.: Language-driven synthesis of {3D} scenes from scene databases. ACM TOG  (2018)

\bibitem{msl11}
Merrell, P., Schkufza, E., Li, Z., Agrawala, M., Koltun, V.: Interactive furniture layout using interior design guidelines. ACM TOG  (2011)

\bibitem{mildenhall2021nerf}
Mildenhall, B., Srinivasan, P.P., Tancik, M., Barron, J.T., Ramamoorthi, R., Ng, R.: {NeRF}: Representing scenes as neural radiance fields for view synthesis. In: ECCV (2020)

\bibitem{openai2023gpt4}
OpenAI: {GPT-4} technical report. arXiv preprint arXiv:2303.08774  (2023)

\bibitem{chatgpt}
Ouyang, L., Wu, J., Jiang, X., Almeida, D., Wainwright, C., Mishkin, P., Zhang, C., Agarwal, S., Slama, K., Ray, A., et~al.: Training language models to follow instructions with human feedback. In: NeurIPS (2022)

\bibitem{cofs}
Para, W.R., Guerrero, P., Mitra, N., Wonka, P.: {COFS}: Controllable furniture layout synthesis. In: SIGGRAPH (2023)

\bibitem{paschalidou2021atiss}
Paschalidou, D., Kar, A., Shugrina, M., Kreis, K., Geiger, A., Fidler, S.: {ATISS}: Autoregressive transformers for indoor scene synthesis. NeurIPS  (2021)

\bibitem{patil2023advances}
Patil, A.G., Patil, S.G., Li, M., Fisher, M., Savva, M., Zhang, H.: Advances in data-driven analysis and synthesis of {3D} indoor scenes. Comput. Graph. Forum  (2023)

\bibitem{qi2017pointnet}
Qi, C.R., Su, H., Mo, K., Guibas, L.J.: {PointNet}: Deep learning on point sets for {3D} classification and segmentation. In: CVPR (2017)

\bibitem{qi2018human}
Qi, S., Zhu, Y., Huang, S., Jiang, C., Zhu, S.C.: Human-centric indoor scene synthesis using stochastic grammar. In: CVPR (2018)

\bibitem{radford2021learning}
Radford, A., Kim, J.W., Hallacy, C., Ramesh, A., Goh, G., Agarwal, S., Sastry, G., Askell, A., Mishkin, P., Clark, J., et~al.: Learning transferable visual models from natural language supervision. In: ICML (2021)

\bibitem{radford2019language}
Radford, A., Wu, J., Child, R., Luan, D., Amodei, D., Sutskever, I.: Language models are unsupervised multitask learners. In: ICML (2019)

\bibitem{rezatofighi2019generalized}
Rezatofighi, H., Tsoi, N., Gwak, J., Sadeghian, A., Reid, I., Savarese, S.: Generalized intersection over union: A metric and a loss for bounding box regression. In: CVPR (2019)

\bibitem{ritchie2019fast}
Ritchie, D., Wang, K., Lin, Y.a.: Fast and flexible indoor scene synthesis via deep convolutional generative models. In: CVPR (2019)

\bibitem{rubenstein2023audiopalm}
Rubenstein, P.K., Asawaroengchai, C., Nguyen, D.D., Bapna, A., Borsos, Z., Quitry, F.d.C., Chen, P., Badawy, D.E., Han, W., Kharitonov, E., et~al.: {AudioPaLM}: A large language model that can speak and listen. arXiv preprint arXiv:2306.12925  (2023)

\bibitem{sch16}
Savva, M., Chang, A.X., Hanrahan, P., Fisher, M., Nie{\ss}ner, M.: {PiGraphs}: learning interaction snapshots from observations. ACM TOG  (2016)

\bibitem{shaw2018self}
Shaw, P., Uszkoreit, J., Vaswani, A.: Self-attention with relative position representations. arXiv preprint arXiv:1803.02155  (2018)

\bibitem{socher2011parsing}
Socher, R., Lin, C.C.Y., Ng, A.Y., Manning, C.D.: Parsing natural scenes and natural language with recursive neural networks. In: ICML (2011)

\bibitem{straub2019replica}
Straub, J., Whelan, T., Ma, L., Chen, Y., Wijmans, E., Green, S., Engel, J.J., Mur-Artal, R., Ren, C., Verma, S., et~al.: The replica dataset: A digital replica of indoor spaces. arXiv preprint arXiv:1906.05797  (2019)

\bibitem{tang2023diffuscene}
Tang, J., Nie, Y., Markhasin, L., Dai, A., Thies, J., Nie{\ss}ner, M.: {DiffuScene}: Scene graph denoising diffusion probabilistic model for generative indoor scene synthesis. arXiv preprint arXiv:2303.14207  (2023)

\bibitem{tang2019learning}
Tang, K., Zhang, H., Wu, B., Luo, W., Liu, W.: Learning to compose dynamic tree structures for visual contexts. In: CVPR (2019)

\bibitem{vaswani2017attention}
Vaswani, A., Shazeer, N., Parmar, N., Uszkoreit, J., Jones, L., Gomez, A.N., Kaiser, {\L}., Polosukhin, I.: Attention is all you need. NeurIPS  (2017)

\bibitem{vinyals2015order}
Vinyals, O., Bengio, S., Kudlur, M.: Order matters: Sequence to sequence for sets. In: ICLR (2016)

\bibitem{wang2019planit}
Wang, K., Lin, Y.A., Weissmann, B., Savva, M., Chang, A.X., Ritchie, D.: {PlanIT}: Planning and instantiating indoor scenes with relation graph and spatial prior networks. ACM TOG  (2019)

\bibitem{wang2021sceneformer}
Wang, X., Yeshwanth, C., Nie{\ss}ner, M.: {SceneFormer}: Indoor scene generation with transformers. In: 3DV (2021)

\bibitem{wei2023lego}
Wei, Q.A., Ding, S., Park, J.J., Sajnani, R., Poulenard, A., Sridhar, S., Guibas, L.: {LEGO-Net}: Learning regular rearrangements of objects in rooms. In: CVPR (2023)

\bibitem{weiss2018fast}
Weiss, T., Litteneker, A., Duncan, N., Nakada, M., Jiang, C., Yu, L.F., Terzopoulos, D.: Fast and scalable position-based layout synthesis. IEEE TVCG  (2018)

\bibitem{xu2018graph2seq}
Xu, K., Wu, L., Wang, Z., Feng, Y., Witbrock, M., Sheinin, V.: Graph2seq: Graph to sequence learning with attention-based neural networks. arXiv preprint arXiv:1804.00823  (2018)

\bibitem{yu2011make}
Yu, L.F., Yeung, S.K., Tang, C.K., Terzopoulos, D., Chan, T.F., Osher, S.J.: {Make It Home}: automatic optimization of furniture arrangement. ACM TOG  (2011)

\bibitem{yyt15}
Yu, L.F., Yeung, S.K., Terzopoulos, D.: The {ClutterPalette}: An interactive tool for detailing indoor scenes. IEEE TVCG  (2015)

\bibitem{deepset}
Zaheer, M., Kottur, S., Ravanbhakhsh, S., P\'{o}czos, B., Salakhutdinov, R., Smola, A.J.: Deep sets. In: NeurIPS (2017)

\bibitem{zellers2018neural}
Zellers, R., Yatskar, M., Thomson, S., Choi, Y.: Neural motifs: Scene graph parsing with global context. In: CVPR (2018)

\bibitem{zhai2023commonscenes}
Zhai, G., {\"O}rnek, E.P., Wu, S.C., Di, Y., Tombari, F., Navab, N., Busam, B.: {CommonScenes}: Generating commonsense {3D} indoor scenes with scene graphs. arXiv preprint arXiv:2305.16283  (2023)

\bibitem{zhao2016relationship}
Zhao, X., Hu, R., Guerrero, P., Mitra, N., Komura, T.: Relationship templates for creating scene variations. ACM TOG  (2016)

\bibitem{Zhao2021Luminous}
Zhao, Y., Lin, K., Jia, Z., Gao, Q.Q., Thattai, G., Thomason, J., Sukhatme, G.: Luminous: Indoor scene generation for embodied ai challenges. In: NeurIPSW (2021)

\bibitem{zhou2019scenegraphnet}
Zhou, Y., While, Z., Kalogerakis, E.: {SceneGraphNet}: Neural message passing for 3{D} indoor scene augmentation. In: CVPR (2019)

\end{thebibliography}

\newpage
\begin{center}
\Large{\textbf{{\model: Revitalizing Order Prior for\\ Sequential Indoor Scene Synthesis}} \\
\vspace{0.5em}{(Supplementary Material)} }
\vspace{1.0em}
\end{center}

\begin{algorithm}[t]
\caption{Pseudocode of Set2Tree algorithm in a PyTorch-like style.}
\label{alg:s2t}
\algcomment{\fontsize{7.2pt}{0em}\selectfont \texttt{iou}: intersection-over-union of two bounding boxes; \texttt{sqrt}: square-root; \texttt{sum}: summarization; \texttt{newaxis}: new axis of an array.
}
\definecolor{codeblue}{rgb}{0.25,0.5,0.5}
\lstset{
  backgroundcolor=\color{white},
  basicstyle=\fontsize{7.2pt}{7.2pt}\ttfamily\selectfont,
  columns=fullflexible,
  breaklines=true,
  captionpos=b,
  commentstyle=\fontsize{7.2pt}{7.2pt}\color{codeblue},
  keywordstyle=\fontsize{7.2pt}{7.2pt},
}
\begin{lstlisting}[language=python]
def giou(boxes1, boxes2):
    # giou between 2 bboxes
    pass
def find_biggest(subscene):
    # find the biggest furniture in a subscene
    pass 
def fn_dist(self, bbox2d_set, xy_set):
    dist_giou = 1 - giou(bbox2d_set) # [N, N]
    dist_enc =  sqrt(sum((xy_set[:, newaxis] - xy_set) ** 2, axis=2)) # [N, N]
    # weighted distance matrix
    return  dist_euc + k * dist_giou

"""
Definition of Furniture: node of Tree
"""
class Furniture:
    def __init__(self, x1y1x2y2=None, xy=None, feature=None):
        self.bbox2d = x1y1x2y2
        self.xy = xy
        self.feature = feature
        self.children = []
    def add_node(self, furniture):
        self.children.append(furniture)
        
"""
Definition of Tree
"""
class Tree:
    def __init__(self):
        self.root = Furniture() # empty 
    def shuffle(self):
        pass # easy to implement
    def bfs(self):
        pass # easy to implement
    def set2tree(self, f_set, labels):
        # organize the furniture set into a scene tree
        for l in len(set(labels)):
            if l != -1: # not outlier
                subscene = [f_set[j] for i in labels if i == l]
                # find the largest furniture as parent node
                f_parent = find_biggest(subscene) 
                # add the parent node under the root node
                self.root.add_node(f_parent)
                # add the remaining objects under the parent node
                for f in subscene:
                    if f != f_parent:
                        f_parent.add_node(f)
             
\end{lstlisting}
\end{algorithm}

\begin{algorithm}[t]
\caption{Pseudocode of Tree2Forest algorithm in a PyTorch-like style.}
\label{alg:t2f}
\vspace{-1mm}
\definecolor{codeblue}{rgb}{0.25,0.5,0.5}
\lstset{
  backgroundcolor=\color{white},
  basicstyle=\fontsize{7.2pt}{7.2pt}\ttfamily\selectfont,
  columns=fullflexible,
  breaklines=true,
  captionpos=b,
  commentstyle=\fontsize{7.2pt}{7.2pt}\color{codeblue},
  keywordstyle=\fontsize{7.2pt}{7.2pt},
}
\begin{lstlisting}[language=python]
from sklearn.cluster import DBSCAN
from numpy import array
from copy import deepcopy
def tree2forest(f_set, eps=0.15, min_samples=2):
    """
    Prepare the distance matrix
    """
    bbox2d_set = array([f.bbox2d for f in f_set]) # [N, 4]
    xy_set = array([f.xy for f in f_set]) # [N, 2]
    dist_mat =  fn_dist(bbox2d_set, xy_set)
    """
    DBSCAN clustering
    """
    dbscan = DBSCAN(eps, min_samples, metric='precomputed')
    labels = dbscan.fit_predict(dist_mat)
    tree = Tree() # empty tree
    tree.set2tree(f_set, labels) # build tree
    forest = [] # create empty forest
    # find all outliner
    outliers = [f_set[j] for j, k in enumerate(labels) if k == -1]
    for f in outliers: # process each outlier
        # add to the parent nodes
        for i in range(len(tree.root.children)):
            tmp_tree = deepcopy(tree)
            tmp_tree.children[i].add_node(f)
            forest.append(tmp_tree)
        # add to the root node
        tmp_tree = deepcopy(tree)
        tmp_tree.root.add_node(f)
        forest.append(tmp_tree)
    return forest
\end{lstlisting}
\end{algorithm}

\begin{algorithm}[t]
\caption{Pseudocode of Forest2Seq algorithm in a PyTorch-like style.}
\label{alg:f2s}
\algcomment{\fontsize{7.2pt}{0em}\selectfont \texttt{bfs}: breadth-first traversal; \texttt{shuffle}: tree shuffle; \texttt{randint}: return random integers from low (inclusive) to high (exclusive).
}
\definecolor{codeblue}{rgb}{0.25,0.5,0.5}
\lstset{
  backgroundcolor=\color{white},
  basicstyle=\fontsize{7.2pt}{7.2pt}\ttfamily\selectfont,
  columns=fullflexible,
  breaklines=true,
  captionpos=b,
  commentstyle=\fontsize{7.2pt}{7.2pt}\color{codeblue},
  keywordstyle=\fontsize{7.2pt}{7.2pt},
}
\begin{lstlisting}[language=python]
"""
Flatten a forest (set of tree) to a sequence
"""
def forest2seq(forest):
    """
    forest: list of Tree
    return: list of Furniture
    """
    N = len(forest) # number of trees
    idx = randint(0, N) # randomly select one tree
    tree_selected = forest[idx]
    return tree_selected.shuffle().bfs()
\end{lstlisting}
\end{algorithm}

This supplementary material contains four sections:
section~\ref{sec:alg} details the algorithm of ordering construction; 
section~\ref{sec:arch} is the architecture details of masked language modeling;
section~\ref{sec:user} provides evaluation details, including the user study and tree reconstruction accuracy; 
section~\ref{sec:vis} offers a comprehensive visual comparison with the-state-of-art methods in various tasks, including scene synthesis, scene completion and rearrangement. 
The source code and pretrained models will also be released upon publication.

\section{Algorithms}
\label{sec:alg}
Our core algorithms can be simply implemented by tens of lines. In training stage, we first use Set2Tree (algorithm~\ref{alg:s2t}, see \textbf{Page 26}) to organize the set of objects into a scene tree, followed by Tree2Forest (algorithm~\ref{alg:t2f}, see \textbf{Page 27}) to expand the single tree to forest representation.
Then we use Forest2Seq (algorithm~\ref{alg:f2s}, see \textbf{Page 27}) to obtain the object sequence from the constructed forest.

\section{Architecture details}
\label{sec:arch}

\noindent\textbf{Layout encoder.}
We employ a ViT~\cite{dosovitskiy2020image}-based layout encoder.
Shown in Table~\ref{tab:layoutenc}, we further conduct ablation study to understand which type of layout encoder is better. We compare our results of the ResNet18~\cite{he2016deep} introduced in ATISS~\cite{paschalidou2021atiss}, PointNet~\cite{qi2017pointnet} employed in LEGO-Net~\cite{wei2023lego}.
We find that with the second-fewest number of parameters, our vit-based layout encoder achieves the best performance.
Additionally, we note that while the PointNet-based layout features primarily concentrate on the shape of the floor plan, they tend to omit scale details.
ResNet18 model tends to disregard low-level features, such as edges, due to its pooling module. However, our vision transformer adapts at capturing these low-level features as evidenced by Figure~\ref{tab:attn}.

\noindent\textbf{Object encoder.}
Following ATISS~\cite{paschalidou2021atiss}, we first apply positional encoding~\cite{vaswani2017attention, mildenhall2021nerf} to each continuous object attribute $h$, including $(r, t, b)$: 
\begin{equation}
    \begin{aligned}
    h' = (\sin{(2^0\pi h)}, \cos{(2^0\pi h)}, 
    \cdots, \\
    \sin{(2^{L-1}\pi h)}, \cos{(2^{L-1}\pi h)}), 
    \end{aligned}
\end{equation}
where $L$=32, $h'\in \mathbb{R}^{64}$ is the encoded scalar and we use a linear projection for $c$ to $c'\in\mathbb{R}^{64}$.
The final object embedding $s\in \mathbb{R}^{512}$ is the concatenation of all attributes. The model is illustrated in Figure~\ref{tab:oe}.

\noindent\textbf{Transformer decoder.}
As illustrated in Figure~\ref{tab:td}, we take GPT2~\cite{radford2019language}-like structure to implement the transformer decoder. Different from the vanilla  transformer~\cite{vaswani2017attention}, it uses layer normalization before self-attention block, adds one additional layer normalization after the final self-attention block, and utilizes GeLU as the activation function.
Our model configuration employs a smaller variant of GPT-2\footnote{https://github.com/karpathy/nanoGPT}, consisting of 6 layers and a 6-head self-attention mechanism, with a hidden dimension size of 192.

\begin{table}[!bt]
	\centering
	\begin{subfigure}{0.6\linewidth}
		\centering
		\includegraphics[width=\linewidth]{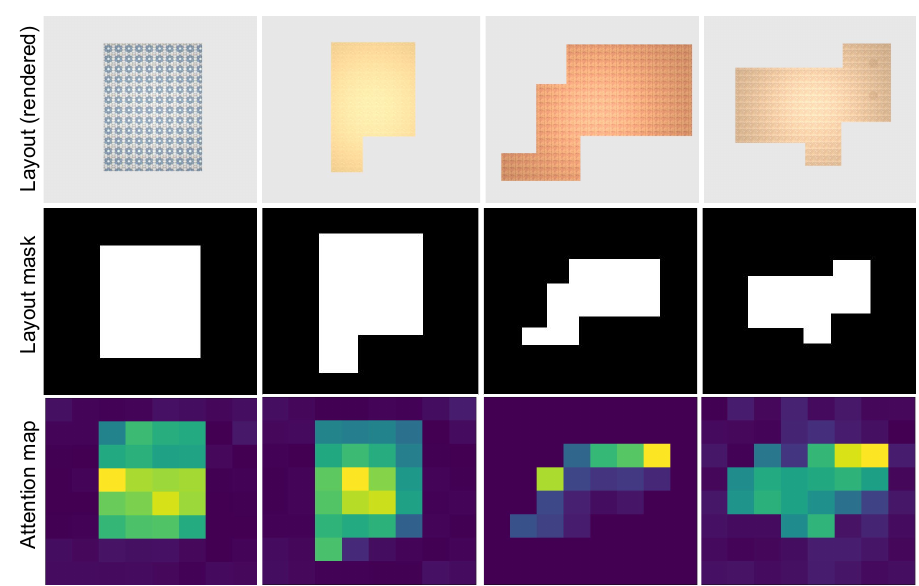}
		\caption{Attention map visualization in the ViT-based layout encoder.}
		\label{tab:attn}
        \hfill
	\end{subfigure}
 	\begin{subfigure}{0.3\linewidth}
		\centering
        \resizebox{\textwidth}{!}{
		\begin{tabular}{l|c|ccc}
\hline
   Model & Parameters (M)  & KL & FID & CAS ($\%$) \\
   \hline 
   PointNet \cite{qi2017pointnet} & \textbf{1.13} &8.3 & 40.5 & 77.4\\
   ResNet18 \cite{he2016deep}  & 11.1  & 6.2 & 36.7  & 70.1\\
   ViT-S \cite{dosovitskiy2020image}  & 2.56  & \textbf{5.9} & \textbf{35.2} & \textbf{68.0} \\
 \hline
\end{tabular}
}
         \vspace{0.57cm}
        \caption{Ablation study on different type of layout encoder.}
        \vspace{0.4cm}
		\label{tab:layoutenc}
	\end{subfigure}

	\vspace{-0.2cm}
	\caption{Ablation study and visualization on the layout encoder.}
\end{table}

\begin{table}[!bt]
	\centering
	\begin{subfigure}{0.33\linewidth}
		\centering
		\includegraphics[width=\linewidth]{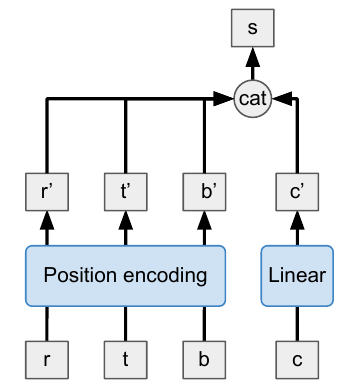}
		\vspace{0.27cm}
		\caption{Object encoder.}
		\label{tab:oe}
	\end{subfigure}
         \hspace{1cm} 
 	\begin{subfigure}{0.26\linewidth}
		\centering
		\includegraphics[width=\linewidth]{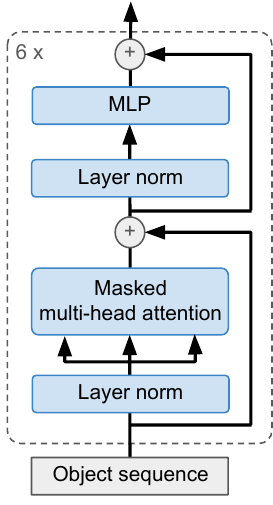}
        \caption{Transformer decoder.} 
		\label{tab:td}
	\end{subfigure}

	\vspace{-0.2cm}
	\caption{Architecture details on the model components.}
\end{table}

\section{Details on evaluation}
\label{sec:user}
\textbf{User study.}
In our user study, each participant evaluates 204 pairs of images (51 samples $\times$ 4 room types). 
They are asked to score the quality of the generated indoor scenes based on two criteria: ``Which scene fits better within the layout boundary?'' and ``Which of these indoor scenes appears more realistic?'', then select the best.

\noindent\textbf{Tree reconstruction accuracy.}
We use Average Hierarchical Distance(AHD) to measure the accuracy of reconstructed tree. A scene tree $\mathcal{T}$ can be decomposed into set of node sets $\{S_i\}_{i=1}^N$, where $N$ is the total depth of the tree and $S_i$ represents all nodes at the $i$-th depth. Further, we compare set similarity in average, formally:
\begin{equation}
    \text{AHD} = \frac{1}{N} \sum_{i=1}^N \frac{|S_i \cap \hat S_i|}{|S_i \cup \hat S_i|},
\end{equation}
where $\hat S_i$ is the set at the $i$-th depth from the ground truth tree. For fair comparison of the different tree reconstruction methods, we select 60 scenes in the experiment and calculate the AHD in average.

\noindent\textbf{Sequence set inconsistency.}
The score is defined as the Hamming distance of two different sequences in average, which can be formulated as:
\begin{equation}
    \text{Inconsistency} = \frac{1}{|N(i, j)|}\sum_{i}\sum_{j\neq i}{d_{\text{Hamming}}(s_i, s_j)},
\end{equation}
where $s_i$ is randomly selected sequence and $N(i, j)$ is the number of $(i, j)$ pairs.
The score is higher when the the sequences within the set are more inconsistent. Specifically, when the set contains only one sequence, the inconsistency score is zero, as there are no pairs of sequences to compare.

\section{Visual results}
\label{sec:vis}
\begin{figure*}
    \centering
    \includegraphics[width=\linewidth]{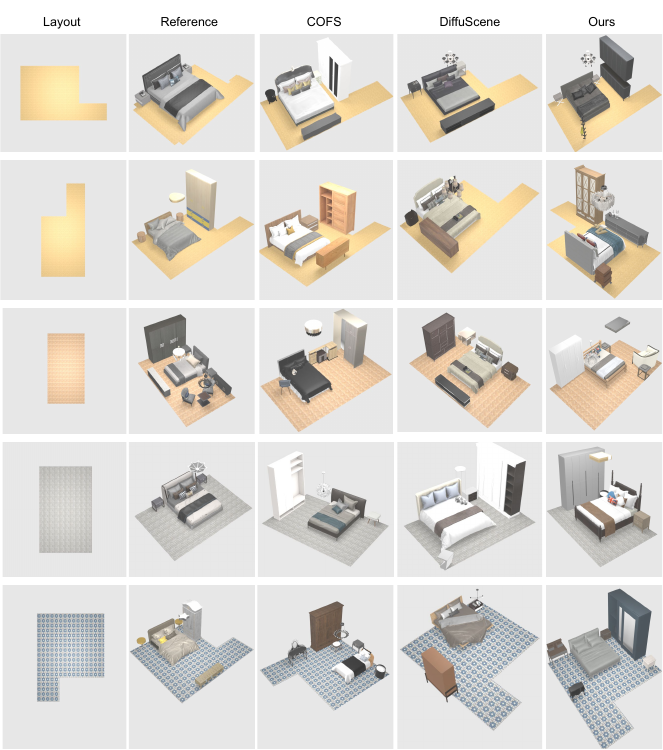}
    \caption{\textbf{Scene synthesis.} Visual comparison with the-state-of-the-art methods~\cite{cofs, tang2023diffuscene} on the bedroom type.}
    \label{fig:supp1}
\end{figure*}

\begin{figure*}
    \centering
    \includegraphics[width=\linewidth]{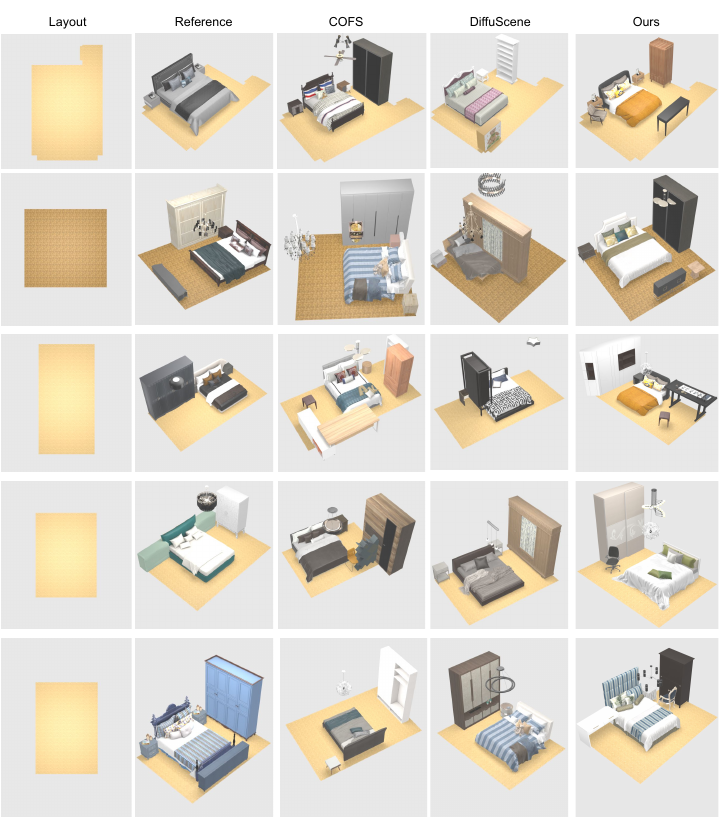}
    \caption{\textbf{Scene synthesis.} More visual comparison with the-state-of-the-art methods~\cite{cofs, tang2023diffuscene} on the bedroom type.}
    \label{fig:supp2}
\end{figure*}

\begin{figure*}
    \centering
    \includegraphics[width=\linewidth]{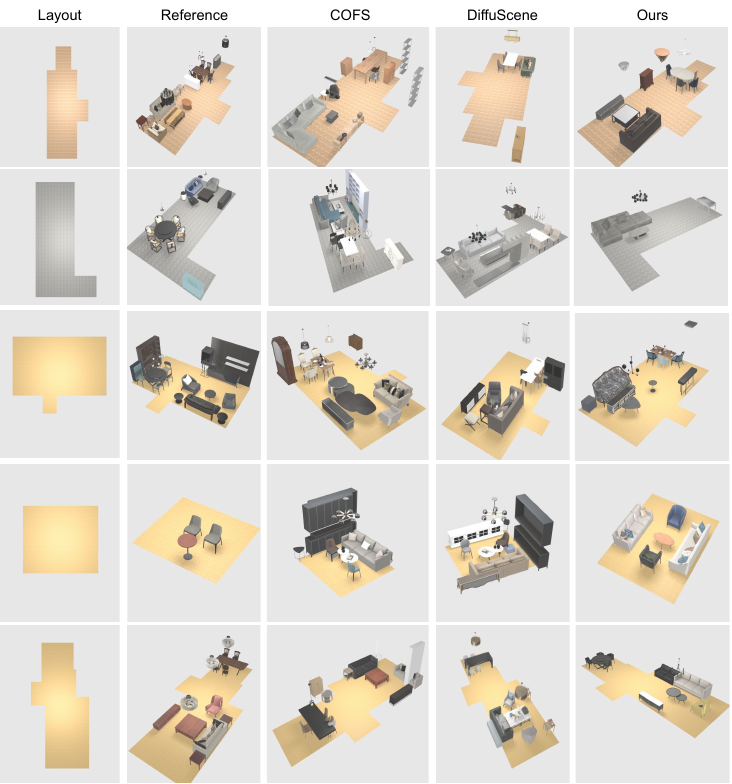}
    \caption{\textbf{Scene synthesis.} Visual comparison with the-state-of-the-art methods~\cite{cofs, tang2023diffuscene} on the living room / dining room type.}
    \label{fig:supp3}
\end{figure*}

\begin{figure*}
    \centering
    \includegraphics[width=\linewidth]{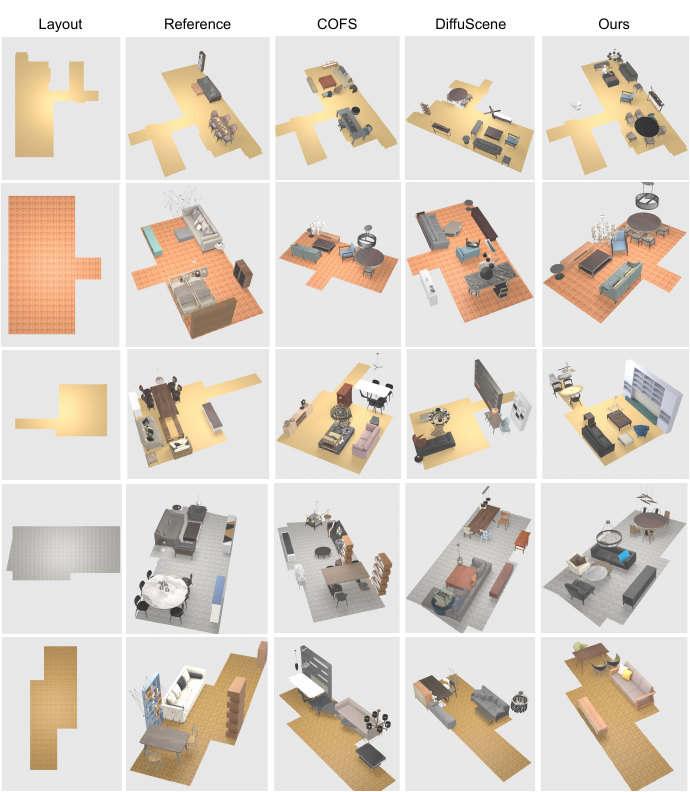}
    \caption{\textbf{Scene synthesis.} More visual comparison with the-state-of-the-art methods~\cite{cofs, tang2023diffuscene} on the living room / dining room type.}
    \label{fig:supp4}
\end{figure*}

\noindent\textbf{More visual comparisons.}
As shown in Figure~\ref{fig:supp1}~\ref{fig:supp2}~\ref{fig:supp3}~\ref{fig:supp4}, we compare our results with the-state-of-the-art methods~\cite{tang2023diffuscene, cofs} in bedroom and living/dining rooms. Our approach yields more realistic scenes with fewer occurrences of objects extending beyond the layout boundaries.

\end{document}